\documentclass{article}

     \PassOptionsToPackage{numbers, compress}{natbib}

    \usepackage[final]{neurips_2019}

\usepackage[utf8]{inputenc} 
\usepackage[T1]{fontenc}    
\usepackage{hyperref}       
\usepackage{url}            
\usepackage{booktabs}       
\usepackage{amsfonts}       
\usepackage{nicefrac}       
\usepackage{microtype}      

\usepackage{graphicx}
\graphicspath{{figures/}}
\usepackage{subcaption}
\usepackage{multirow}

\title{Sensor Fusion using Backward Shortcut Connections for Sleep Apnea Detection in Multi-Modal Data}

\author{%
  Tom Van Steenkiste, Dirk Deschrijver, Tom Dhaene\\
  Ghent university - imec, IDLab\\
  Technologiepark-zwijnaarde 126\\
  9052 Gent, Belgium \\
  \texttt{tomd.vansteenkiste@ugent.be} \\
}

\begin{document}

\maketitle

\begin{abstract}
Sleep apnea is a common respiratory disorder characterized by breathing pauses during the night. Consequences of untreated sleep apnea can be severe. Still, many people remain undiagnosed due to shortages of hospital beds and trained sleep technicians. To assist in the diagnosis process, automated detection methods are being developed. Recent works have demonstrated that deep learning models can extract useful information from raw respiratory data and that such models can be used as a robust sleep apnea detector. However, trained sleep technicians take into account multiple sensor signals when annotating sleep recordings instead of relying on a single respiratory estimate. To improve the predictive performance and reliability of the models, early and late sensor fusion methods are explored in this work. In addition, a novel late sensor fusion method is proposed which uses backward shortcut connections to improve the learning of the first stages of the models. The performance of these fusion methods is analyzed using CNN as well as LSTM deep learning base-models. The results demonstrate a significant and consistent improvement in predictive performance over the single sensor methods and over the other explored sensor fusion methods, by using the proposed sensor fusion method with backward shortcut connections.

\end{abstract}

\section{Introduction}
Sleep apnea is a respiratory disorder consisting of breathing pauses, or apneaic events, during the night~\cite{guilleminault1976sleep}. These events are either classified as obstructive sleep apnea (OSA) when the upper airway collapses or central sleep apnea (CSA) when the signals to control the breathing are disturbed. When breathing becomes shallow, but is not yet fully disturbed, it is classified as hypopnea.

Consequences of untreated sleep apnea can be severe ranging from hypertension, cardiac arrhythmia to even strokes and heart failure~\cite{somers2008sleep,yaggi2005obstructive}. It is estimated that 49.7\% of male and 23.4\% of female adults suffer from some form of sleeping related breathing disorder~\cite{heinzer2015prevalence}. However, many of these patients are unaware of their condition and suspected patients are faced with long waiting times for diagnosis due to expensive setups and a limited amount of hospital beds and trained sleep technicians. In the USA, waiting times of up to 60 months have been reported~\cite{flemons2004access}. 

Before diagnosis, patients are typically admitted for an overnight sleep study using a polysomnography (PSG)~\cite{berry2016aasm}. This device measures a wide range of sensor data including respiration, heart rate, oxygen saturation and brain activity. After the recording, trained sleep technicians inspect the signals of the entire night and manually annotate the data for sleep events, including sleep apnea episodes, using a standard reference such as the AASM guidelines~\cite{berry2016aasm}. The severity of sleep apnea is then summarized using the Apnea-Hypopnea-Index (AHI) which represents the number of apnea or hypopnea events per hour of sleep.

To assist in this process, and reduce the load on personnel, automated scoring methods have been investigated. These methods range from rule-based algorithms to machine learning approaches. The machine learning models generally use human-engineered features. However, thanks to advancements in deep learning technology, researchers have been able to use deep learning models for extracting features and detecting sleep apnea events in raw respiratory data. Examples include a CNN based model proposed by Haidar et al.~\cite{haidar2017sleep} which uses the nasal airflow signal and an LSTM based model proposed by Van Steenkiste et al.~\cite{van2018automated} which was tested on the abdominal respiratory belt, thoracic respiratory belt and the ECG derived respiration signal. 

Although these deep learning methods have offered an increase in detection performance and robustness, a crucial limitation is their inclusion of only a single respiratory channel whereas sleep technicians take into account a wide range of sensor data when analyzing the sleep of a patient. In addition to respiration, other important parameters include the heart rate~\cite{snyder1964changes} and oxygen saturation~\cite{berry2016aasm}. To further improve the performance and reliability of these deep learning models, it is crucial that data from the other recorded signals is included into the detection model.

Combining data from multiple sources is known as sensor fusion and has been successfully applied across a wide range of medical use-cases including activity recognition~\cite{gravina2017multi,munzner2017cnn}, sleep quality analysis~\cite{peng2007multimodality} and sleep detection~\cite{chen2017multimodal}. When the data is combined in the beginning of the model, the method is denoted as early fusion. When, on the other hand, the data is combined at a later stage, the method is known as late fusion. In deep learning, the early fusion method consists of a multi-input model where all data is simply concatenated. The late fusion method consists of separate branch networks for each input, that are combined at a later stage.

In this work, the performance of these two sensor fusion approaches is analyzed and a novel late fusion method using backward shortcut connections is proposed for combining data from the abdominal respiratory belt, thoracic respiratory belt, heart rate and oxygen saturation into a single deep learning model. The performance and robustness is verified using a CNN~\cite{lecun1990handwritten} as well as an LSTM~\cite{hochreiter1997long} based model.

In Section~\ref{sec:baseline}, the base single-input deep learning models are presented. Then, in Section~\ref{sec:sensorfusion}, the various sensor fusion methods are discussed and the novel approach is presented. In Section~\ref{sec:experimentalsetup}, the experimental setup is detailed and in Section~\ref{sec:resultsanddiscussion} the results of the experiments are presented and discussed. Finally, conclusions are drawn in Section~\ref{sec:conclusion}.

\section{Single Input Methods}
\label{sec:baseline}

\begin{figure}[h]
	\centering
	\begin{subfigure}[h]{0.59\textwidth}
		\includegraphics[width=\textwidth]{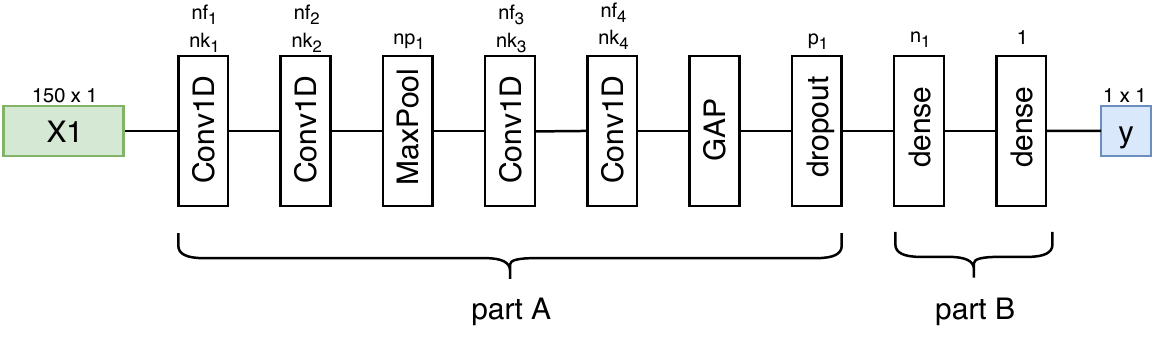}
		\caption{CNN-based model.}
		\label{fig:simcnn}
	\end{subfigure}
	
	\begin{subfigure}[h]{0.49\textwidth}
		\includegraphics[width=\textwidth]{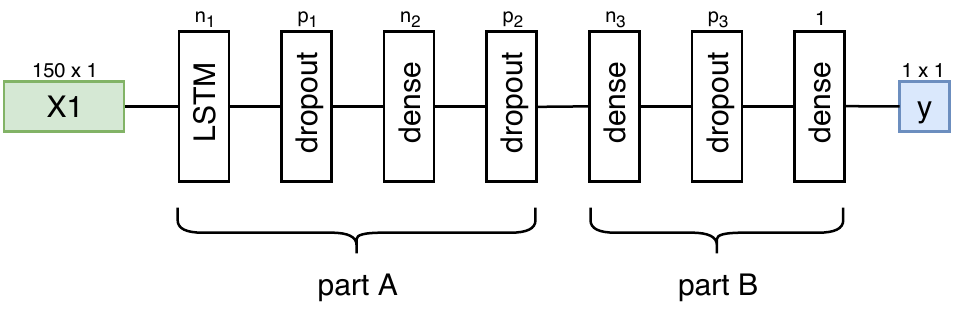}
		\caption{LSTM-based model.}
		\label{fig:simlstm}
	\end{subfigure}
	\caption{State-of-the-art baseline CNN and LSTM models for the automated detection of sleep apnea events in raw respiratory data. The models are split into two parts to be used by the sensor fusion methods. Here, $X1$ represents the raw respiratory data input and $y$ represents the binary sleep apnea prediction label.}
	\label{fig:simmodels}
\end{figure}

Several deep-learning architectures for automated sleep apnea detection have been proposed. In \cite{haidar2017sleep}, Haidar et al. suggest a CNN based approach using data from the nasal airflow sensor. In \cite{van2018automated}, 
Van Steenkiste et al. suggest an LSTM based approach which was validated for data from the abdominal respiratory belt, thoracic respiratory belt and ECG derived respiration. However, data from these multiple sensors has not yet been combined into a single model. Based on these works, two baseline deep learning models are defined as shown in Figure~\ref{fig:simmodels}. 

The CNN-based model in Figure~\ref{fig:simcnn} consists of two convolutional layers~\cite{lecun1990handwritten} followed by a maxpooling layer, two more convolutional layers and a Global Average Pooling (GAP) layer. A dropout layer is added for regularization~\cite{srivastava2014dropout} and two dense layers provide the final predictions of the model. All layers have \textit{relu} activations except for the final dense layer which has a \textit{sigmoid} activation for binary prediction. The number of filters in the first two convolutional layers $nf_1$ and $nf_2$ is 100 whereas the number of filters in the final two convolutional layers $nf_3$ and $nf_4$ is 160. All convolutional layers have a kernel size $nk_1$, $nk_2$, $nk_3$ and $nk_4$ of 10. The max pooling layer has a pool size of $np_1$ of 3 and the dropout probability $p_1$ is 50\%. 

The LSTM-based model in Figure~\ref{fig:simlstm} consists of an LSTM layer~\cite{hochreiter1997long} followed by three groups of dropout and dense layers. The number of LSTM nodes $n_1$ is 50 and their activation function is the \textit{tanh} function. The first two dense layers consist of $n_2$ and $n_3$ number of nodes equal to 25 with a \textit{relu} activation function. The final dense node has a \textit{sigmoid} activation function for binary prediction. All dropout probabilities $p_1$, $p_2$ and $p_3$ equal 20\%.

\section{Sensor Fusion Methods}
\label{sec:sensorfusion}
\begin{figure}[h]
	\centering
	\begin{subfigure}[h]{0.27\textwidth}
		\centering
		\includegraphics[width=\textwidth]{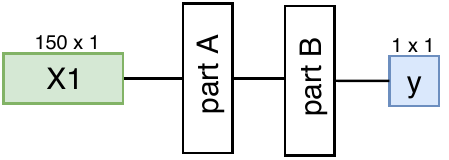}
		\caption{Single Input Method (SIM).}
		\label{fig:SIM}
	\end{subfigure}
	~
	\begin{subfigure}[h]{0.25\textwidth}
		\centering
		\includegraphics[width=\textwidth]{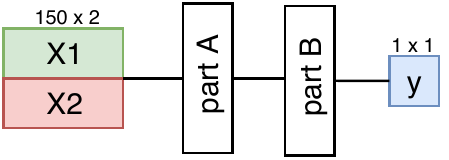}
		\caption{Multi Input Method (MIM).}
		\label{fig:MIM}
	\end{subfigure}
	
	\begin{subfigure}[h]{0.28\textwidth}
		\centering
		\includegraphics[width=\textwidth]{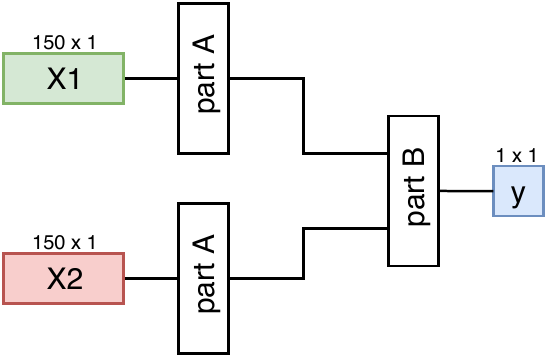}
		\caption{Branched Fusion Method (BFM).}
		\label{fig:BFM}
	\end{subfigure}
	~
	\begin{subfigure}[h]{0.30\textwidth}
		\centering
		\includegraphics[width=\textwidth]{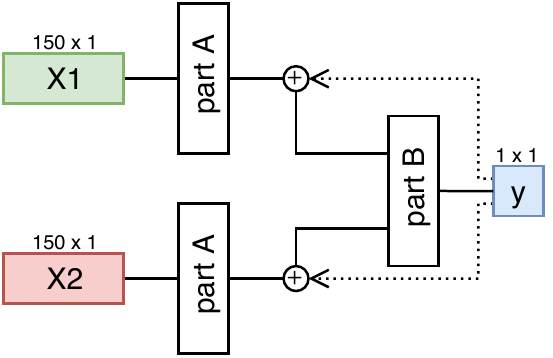}
		\caption{Branched Fusion Method with Backward Shortcut Connections (BFM-SC).}
		\label{fig:BFMSC}
	\end{subfigure}
	
	\caption{Examples of different types of methods for fusion of multiple sensors. In the examples, data from two sensors is used.}
	\label{fig:fusion}
\end{figure}

Sensor fusion refers to the combination of data from multiple sensors into one single decision model. This can be achieved at three different levels: The data level, the feature level and the decision level~\cite{gravina2017multi,liggins2017handbook}. In deep learning, features can be automatically learned and extracted from raw data. In early fusion models, the data is passed into the model as a single input matrix by concatenating all input data. In late fusion models, the deep learning architecture is split up into several branches. The different possible configurations are shown in Figure~\ref{fig:fusion}.


Figure~\ref{fig:SIM} represents the Single Input Method (SIM) as presented in Section~\ref{sec:baseline}. The original deep learning model is split up into a \textit{part A} representing the first stages and a \textit{part B} representing the later stages as demonstrated in Figure~\ref{fig:simmodels}. The SIM can only use a single input signal. However, the method can easily be extended to include multiple inputs as shown in Figure~\ref{fig:MIM}. This method is denoted as the Multi Input Method (MIM) and consists of the same architecture as the SIM but with a concatenated input matrix. Using this early fusion approach, the model is able to learn features that are correlated across the sensors. However, there is no intrinsic incentive for the model to extract as much information as possible form each input sensor individually as training and optimization procedures only focus on the combined prediction. The MIM has, for example, been applied for activity recognition~\cite{munzner2017cnn} based on CNN networks and sleep detection~\cite{chen2017multimodal} based on LSTM networks using human-engineered features.

In Figure~\ref{fig:BFM}, the deep learning architecture is split up into several branches for what is denoted as the Branched Fusion Method (BFM). In this late fusion setup, correlations are being learned at the feature level instead of at the data level. This type of fusion model is often used to fuse data from multiple streams in deep learning architectures. It has, for example, been used in activity recognition~\cite{munzner2017cnn} based on CNN networks, sleep quality monitoring~\cite{peng2007multimodality} based on SVM models, and driver activity anticipation~\cite{jain2016recurrent} and medical procedure monitoring~\cite{bernal2018deep} both based on LSTM networks.

A disadvantage of the BFM is that the separate branches often do not reach their full potential. There is no incentive for each of the branches to perform at their full potential. This can be derived by inspecting the error function used during the backpropagation step as defined in Equation~\ref{eq:errorbfm} where $L_{B_i}$ represents the final layer of branch $i$, $L_f$ represents the first layer of the fusion part, $w^{L_f,L_{B_i}}$ represents the weights between the branch and the fusion part, $z^{L_{B_i}}$ represents the weighted input to branch $B_i$ and $\sigma$ is the activation function.

\begin{equation}
\label{eq:errorbfm}
\delta^{L_{B_i}}_{BFM} = ((w^{L_f,L_{B_i}})^T \delta^{L_f}) \odot \sigma'(z^{L_{B_i}})
\end{equation}

When one of the branches is able to capture the required behavior, the $\delta_o$ at the output node will be near zero. This will be propagated via $\delta^{L_f}$ to $\delta^{L_{B_i}}$ for each branch $i$, resulting in small updates to the branches and the network in general. For the branch or branches influencing this output the most, this is to be expected and desired as the output is close to the target value. For the other branches, however, this can critically limit the learning capacity as they are only subjected to small weight updates from that point forward. The first branch or set of branches to accurately predict the events will lead the model and the potential benefit of including the other sensors is lost.

\begin{figure}[!htb]
	\centering
	\begin{subfigure}[h]{0.44\textwidth}
		\includegraphics[width=\textwidth]{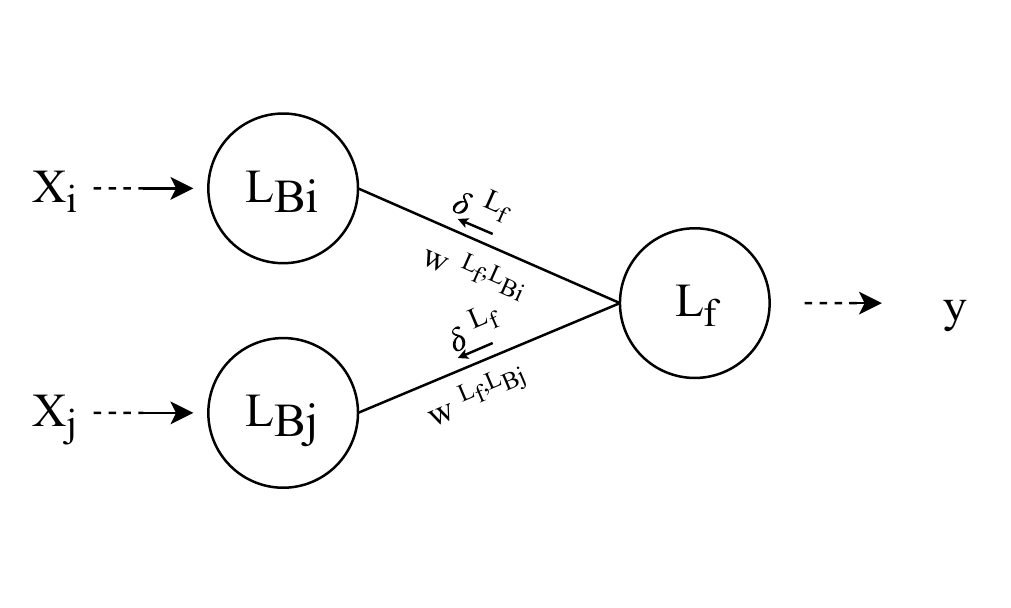}
		\caption{BFM.}
		\label{fig:errorbfm}
	\end{subfigure}
	~
	\begin{subfigure}[h]{0.44\textwidth}
		\includegraphics[width=\textwidth]{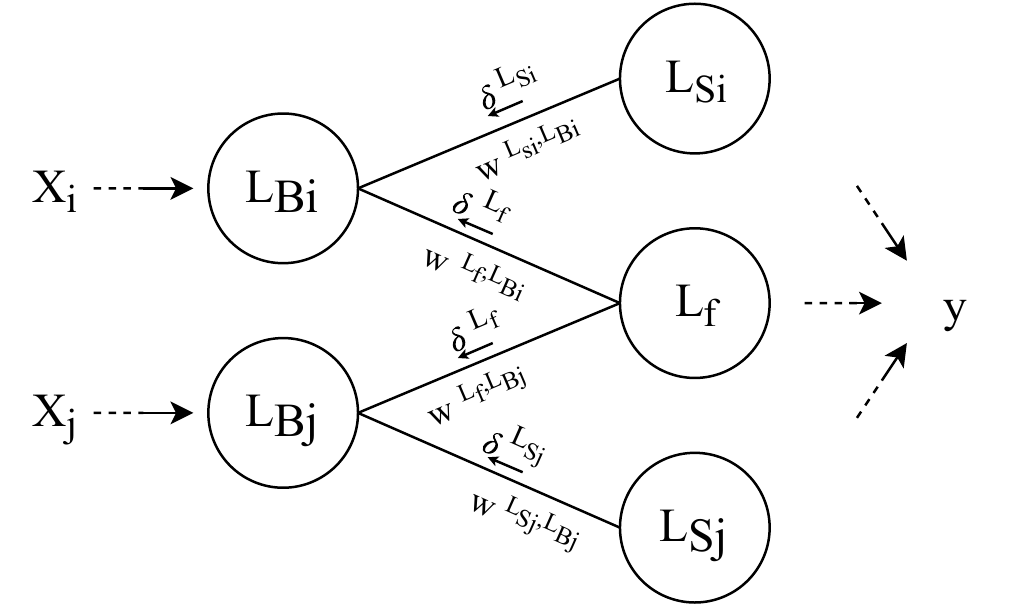}
		\caption{BFM-SC.}
		\label{fig:errorbfmsc}
	\end{subfigure}
	\caption{Error propagation in the BFM and BFM-SC. Here, $L_{B_i}$ is the final layer of the $i$th branch, $L_{f}$ is the first layer of the fusion part and $L_{S_i}$ is the first layer of the shortcut connection for branch $i$.}
	\label{fig:errors}
\end{figure}

In Figure~\ref{fig:BFMSC}, a novel approach is proposed which uses backward shortcut connections to enhance the learning capacity of each separate branch: the Branched Fusion Method with Shortcut Connections (BFM-SC). Shortcut connections have been used before, e.g. as skip connections in residual neural networks~\cite{he2016deep}. In that setting, the connections are used to pass the input forward to the next layer. Here, they are used to pass the target back to the different branches. This incentivizes each part of the model, i.e. the separate branches and the fusion layer, to work at maximum performance. The backward shortcut connections can be implemented by adding a separate \textit{part B} for each input branch, each with its own output. The branch thus provides input to its own \textit{part B} and the fusion \textit{part B}. This results in two different error measures being propagated back to the branch as visualized in Figure~\ref{fig:errors}. The error $\delta^{L_{B_i}}$ for branch $i$ then consists of the combination of the shortcut connection error $\delta^{L_{S_i}}$ and the original fusion part error $\delta^{L_f}$ as demonstrated in Equation~\ref{eq:errorbfmsc}.

\begin{equation}
\label{eq:errorbfmsc}
\delta^{L_{B_i}}_{BFM-SC} = ((w^{L_f,L_{B_i}})^T \delta^{L_f}) \odot \sigma'(z^{L_{B_i}}) / 2 + ((w^{L_{S_i},L_{B_i}})^T \delta^{L_{S_i}}) \odot \sigma'(z^{L_{B_i}}) / 2
\end{equation}

When one or more of the branches of the BFM-SC take the lead, resulting in a small $\delta_o$ and thus a small $\delta^{L_f}$, the other branches still get weight updates via their respective $\delta^{L_{S_i}}$ and hence they can continue learning. The output of these branches will continue to change, causing updates to the weights $w^{L_f,L_{B_i}}$ connecting the branch to the fusion part. Consequently, the importance of the branch in the fusion output continuously changes while the performance is increasing and branches that take longer to reach their full performance can still be included into the model. When the BFM-SC is deployed and used, the additions of to the backward shortcut connections can be removed and the model functions as a regular BFM.

\section{Experimental Setup}
\label{sec:experimentalsetup}
To analyze and compare the performance of the different methods, data from the Sleep-Heart-Health-Study-1 database~\cite{quan1997sleep} is used which contains PSG recordings of 5804 adults of age 40 and older. From this database, five non-overlapping datasets, or groups, of 100 patients each are created. In order to accurately assess the performance for sleep apnea detection in hospital settings, there are no specific inclusion criteria and patients are simply sequentially included. Hence, the first group of patients corresponds to the first 100 patients in the SHHS-1 database. Each of the groups is split in 30 patients for training, 20 patients for validation and 50 patients for testing. An overview of patient characteristics is provided in Table~\ref{table:meta}. For privacy, age is represented as a categorical variable with five representing patients between 35 and 44 years of age and ten representing patients 85 year or older. Each step represents an increase of 10 years.

\begin{table}[h]
	\centering
	\caption{Patient characteristics for the different groups used in this study. The AHI metric demonstrates that group 2 has the most sleep apnea events per hour of sleep. \label{table:meta}}
	\begin{tabular}{@{}ccccc@{}}
		\toprule
		group & AHI [events/h]             & BMI [kg/$m^2$]            & gender [\% male]      & age [category]           \\ \midrule
		0     & $12.8 \pm 13.5$ & $27.0 \pm 4.4$ & $37$ & $7.1 \pm 1.1$ \\
		1     & $13.3 \pm 16.4$ & $27.5 \pm 5.7$ & $56$ & $6.7 \pm 1.3$ \\
		2     & $19.6 \pm 17.3$ & $27.3 \pm 4.4$ & $40$ & $7.2 \pm 1.1$ \\
		3     & $13.7 \pm 12.2$ & $27.2 \pm 4.9$ & $52$ & $7.0 \pm 1.3$ \\
		4     & $10.9 \pm 10.0$ & $26.8 \pm 4.7$ & $55$ & $7.3 \pm 1.3$ \\ \bottomrule
	\end{tabular}
\end{table}

For each patient, the abdominal respiratory belt (abdores), thoracic respiratory belt (thorres), heart rate (HR) and oxygen saturation (SaO2) are extracted in addition to the human-based annotations of OSA, CSA and hypopnea. All recorded signals are resampled to 5Hz. To reduce noise and enable the deep learning models to focus on the relevant aspects of the signals, the data is passed through a basic preprocessing stage. The respiratory belt signals are filtered using a fourth-order low-pass zero-phase-shift Butterworth filter with a cut-off frequency of 0.7Hz to extract relevant respiratory information~\cite{hettrick2006bioimpedance}. Next, they are normalized using the 5\% and 95\% percentile per patient. Sensor artifacts in the HR and SaO2 signals are removed through linear interpolation of missing values. The HR signal is normalized based on a minimum and maximum heart rate of 50 and 105 beats per minute respectively. The SaO2 signal is normalized between an 80\% and 100\% saturation interval. The three annotation signals OSA, CSA and hypopnea are combined into a single binary indicator signal.

The preprocessed data are split up into epochs of 30 seconds with a stride of 1 second. If, at any point during the epoch, there was a human-based annotation of apnea, the entire epoch is labeled as apnea-positive. Apnea epochs are less common than normal sleep leading to an imbalanced data problem. For training, the dataset is balanced via random balanced sampling. For testing and validation, the dataset is not altered to accurately represent the real-life problem case.

The generated datasets are used to benchmark the performance of the different methods presented in Section~\ref{sec:sensorfusion}. Each of these methods is tested using a CNN-based model as well as using an LSTM-based model. For each of the four input signals, a separate CNN and LSTM model with the SIM is constructed. The four signals are also combined to generate a CNN and LSTM model with the MIM, BFM and BFM-SC. This results in 14 different model configurations being analyzed. Each configuration is trained using the binary-crossentropy loss function with an Adam optimizer~\cite{kingma2014adam} with learning rate set to $0.001$ for 25 epochs. Gradients are clipped at $0.5$ to improve learning. Early stopping based on the validation set is applied~\cite{morgan1990generalization,caruana2001overfitting}. 

To accurately assess the performance, the different model configurations are trained and evaluated five times using a different group of patients. In addition, the robustness of the configurations is assessed by training and evaluating the models an additional four times on the first group of patients. This results in a total of nine experiments for each configuration.

The primary endpoint of this study is the performance of the model for detecting sleep apnea events. As these experiment deal with an unbalanced dataset, the area under the precision-recall curve (AUPR) is used as is recommended in literature~\cite{kotsiantis2006handling,he2009learning}. To test the statistical significance of any improvement of the BFM-SC over the other configurations, a paired T-test is used. In addition, the robustness of the improvement of the BFM-SC over the different other configurations is analyzed using a heteroscedastic T-test over the five repetitions on the first group of patients. For each experiment, a significance level of $p<0.05$ is required.

\section{Results and Discussion}
\label{sec:resultsanddiscussion}
\begin{figure}[!htb]
	\centering
	\begin{subfigure}[h]{0.44\textwidth}
		\includegraphics[width=\textwidth]{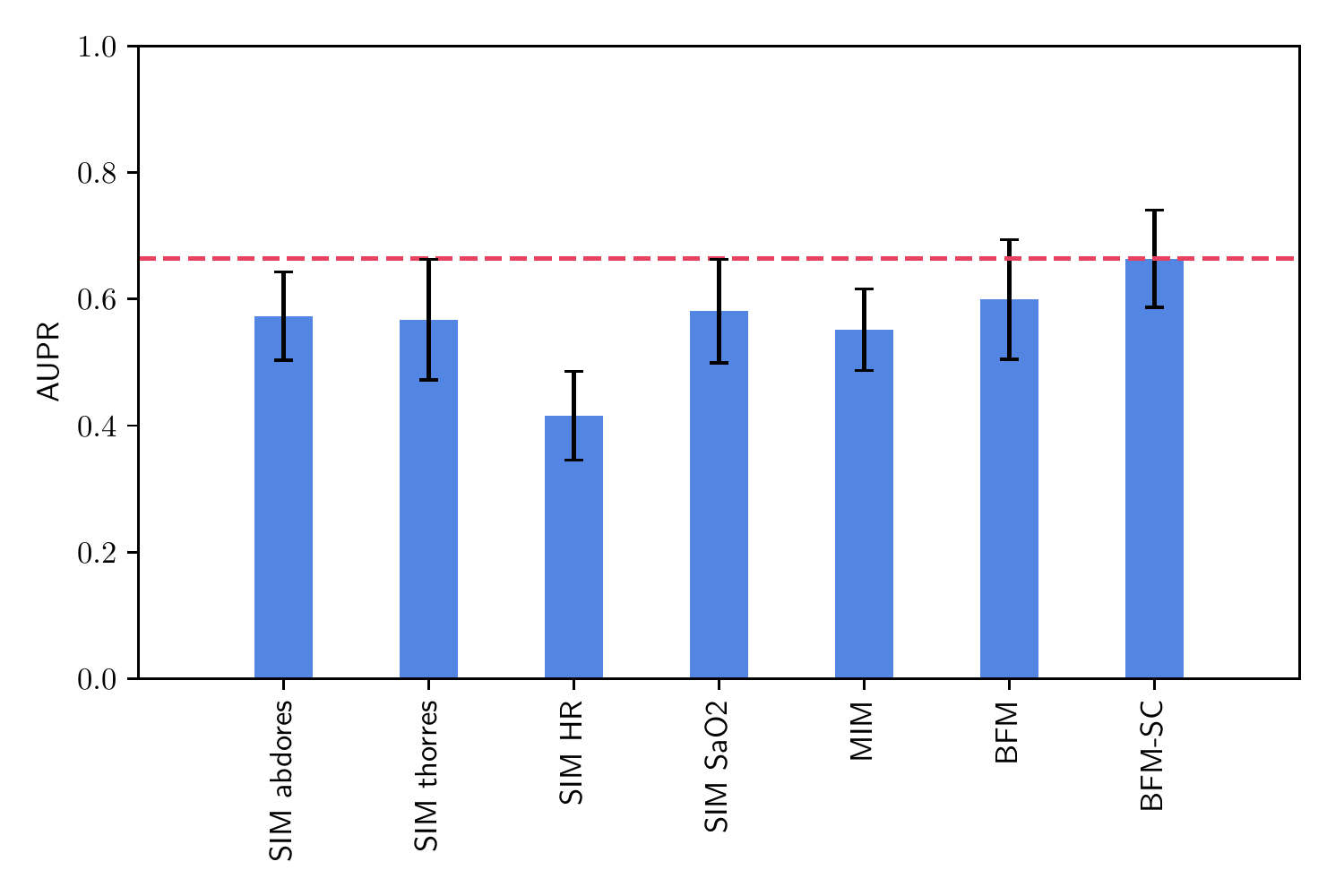}
		\caption{CNN-based model.}
		\label{fig:}
	\end{subfigure}
	~
	\begin{subfigure}[h]{0.44\textwidth}
		\includegraphics[width=\textwidth]{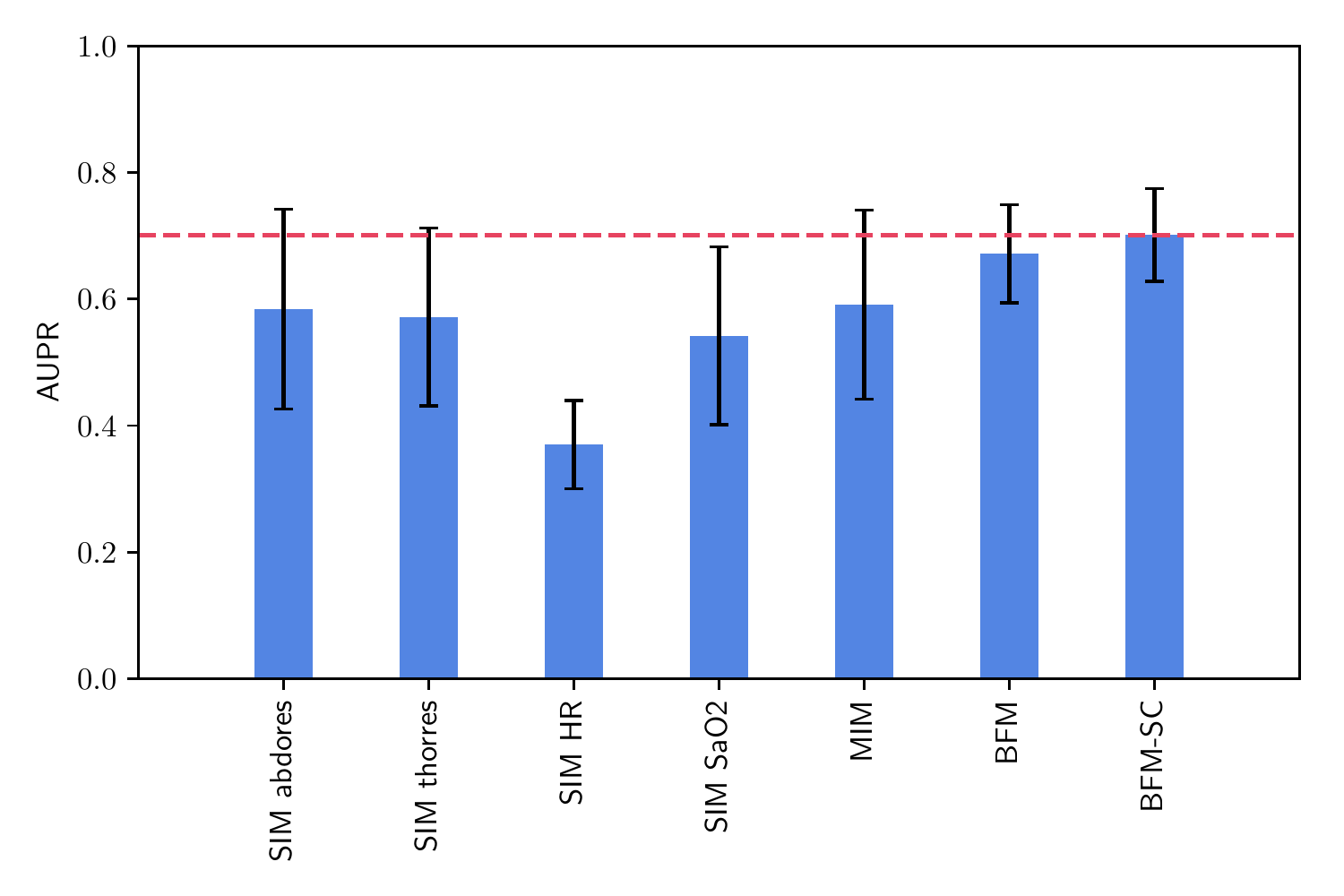}
		\caption{LSTM-based model.}
		\label{fig:}
	\end{subfigure}
	\caption{Summary of the performance of the different models across the different datasets. The BFM-SC approach results in an improvement over all other methods. The red line indicates the mean score of the BFM-SC.}
	\label{fig:diff}
\end{figure}

All experiments showed convergence before the 25 epoch threshold. The results of the five runs to test the performance of the models are shown in Table~\ref{table:cnndiff} and Table~\ref{table:lstmdiff} for the CNN-based and LSTM-based configurations respectively. These results demonstrate that the BFM-SC approach consistently performs better than the other approaches. Figure~\ref{fig:diff} summarizes the performance of the different approaches.

\begin{table}[!htb]
	\centering
	\caption{AUPR scores for the different tested approaches using the CNN-based model. The BFM-SC outperforms all other configurations.\label{table:cnndiff}}
	\begin{tabular}{@{}llllll@{}}
		\toprule
		& group 0 & group 1 & group 2 & group 3 & group 4 \\ \midrule
		SIM abdores & 0.70    & 0.59    & 0.55    & 0.51    & 0.51    \\
		SIM thorres & 0.64    & 0.55    & 0.69    & 0.54    & 0.41    \\
		SIM hr      & 0.44    & 0.35    & 0.54    & 0.37    & 0.37    \\
		SIM SaO2    & 0.63    & 0.58    & 0.70    & 0.53    & 0.47    \\
		MIM         & 0.62    & 0.58    & 0.61    & 0.48    & 0.47    \\
		BFM         & 0.67    & 0.66    & 0.69    & 0.53    & 0.45    \\
		BFM-SC      & \textbf{0.74}    & \textbf{0.68}    & \textbf{0.75}    & \textbf{0.57}    & \textbf{0.58}    \\ \bottomrule
	\end{tabular}
\end{table}

\begin{table}[!htb]
	\centering
	\caption{AUPR scores for the different tested approaches using the LSTM-based model. The BFM-SC outperforms all other configurations.\label{table:lstmdiff}}
	\begin{tabular}{@{}lccccc@{}}
		\toprule
		& group 0 & group 1 & group 2 & group 3 & group 4 \\ \midrule
		SIM abdores & 0.73    & 0.68    & 0.65    & 0.57    & 0.29    \\
		SIM thorres & 0.71    & 0.53    & 0.73    & 0.54    & 0.34    \\
		SIM hr      & 0.36    & 0.37    & 0.50    & 0.29    & 0.33    \\
		SIM SaO2    & 0.65    & 0.57    & 0.71    & 0.31    & 0.46    \\
		MIM         & 0.72    & 0.63    & 0.72    & 0.57    & 0.31    \\
		BFM         & 0.76    & 0.67    & 0.76    & 0.61    & 0.56    \\
		BFM-SC      & \textbf{0.78}    & \textbf{0.70}    & \textbf{0.78}    & \textbf{0.64}    & \textbf{0.61}    \\ \bottomrule
	\end{tabular}
\end{table}

The variation in performance across the groups can be explained by the sensitivity of the AUPR metric to the positive-negative balance in the labels. For an accurate estimation of the performance, these scores need to be compared against the performance of a random binomial model with predictions based on the balance of positive and negative events in the training data. These scores are presented in Table~\ref{table:random} for each of the groups. As would be expected, there is a strong correlation with the AHI values of each group presented in Table~\ref{table:meta}.

\begin{table}[!htb]
	\centering
	\caption{AUPR scores for a random binomial prediction model based on the intrinsic imbalance in each training data set. The difference in balance influences the baseline performance for each group.\label{table:random}}
	\begin{tabular}{@{}ccccc@{}}
		\toprule
		group 0    & group 1     & group 2   & group 3     & group 4     \\ \midrule
		0.36 & 0.36 & 0.42 & 0.31 & 0.29 \\ \bottomrule
	\end{tabular}
\end{table}

\begin{figure}[!htb]
	\centering
	\begin{subfigure}[h]{0.44\textwidth}
		\includegraphics[width=\textwidth]{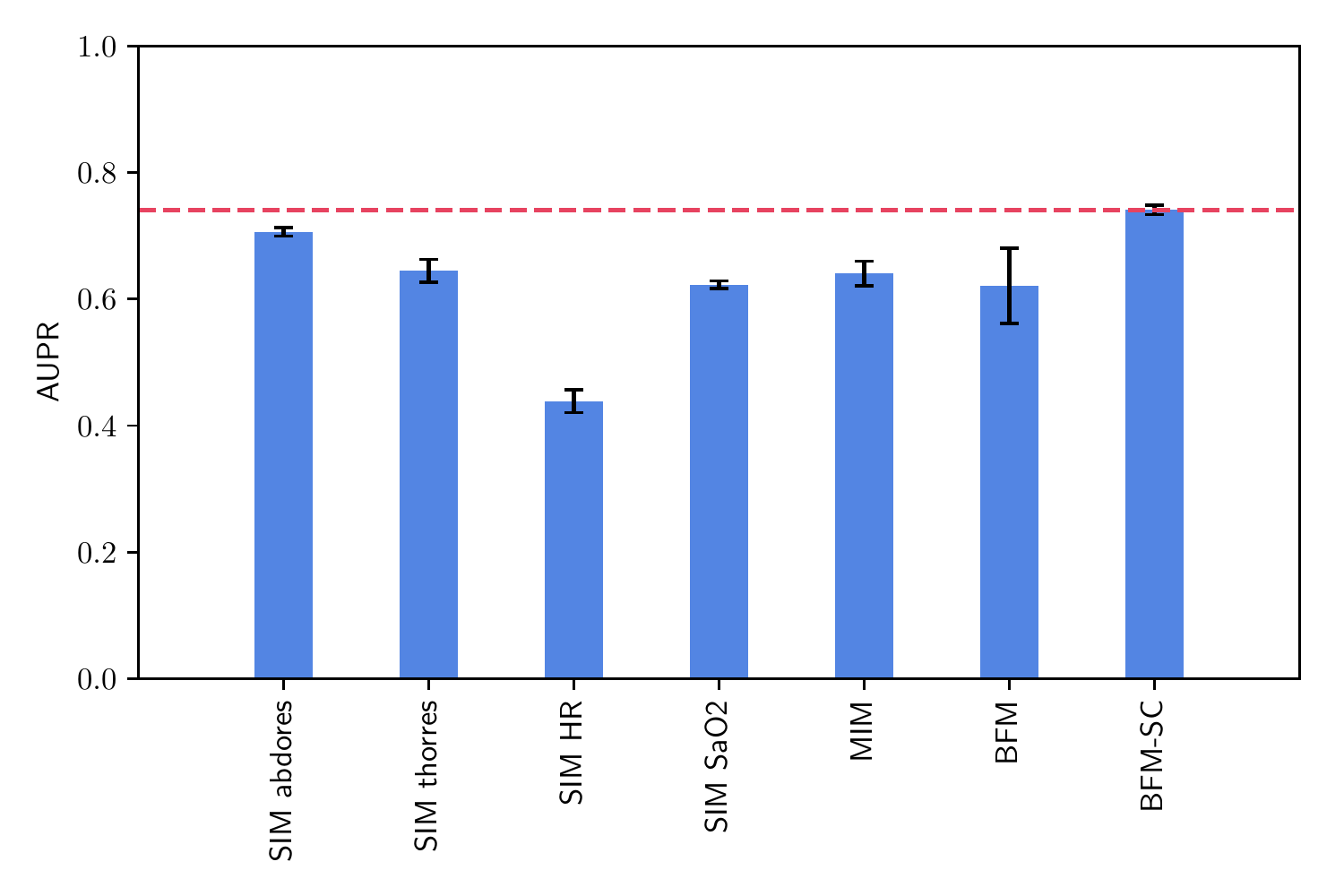}
		\caption{CNN-based model.}
		\label{fig:}
	\end{subfigure}
	~
	\begin{subfigure}[h]{0.44\textwidth}
		\includegraphics[width=\textwidth]{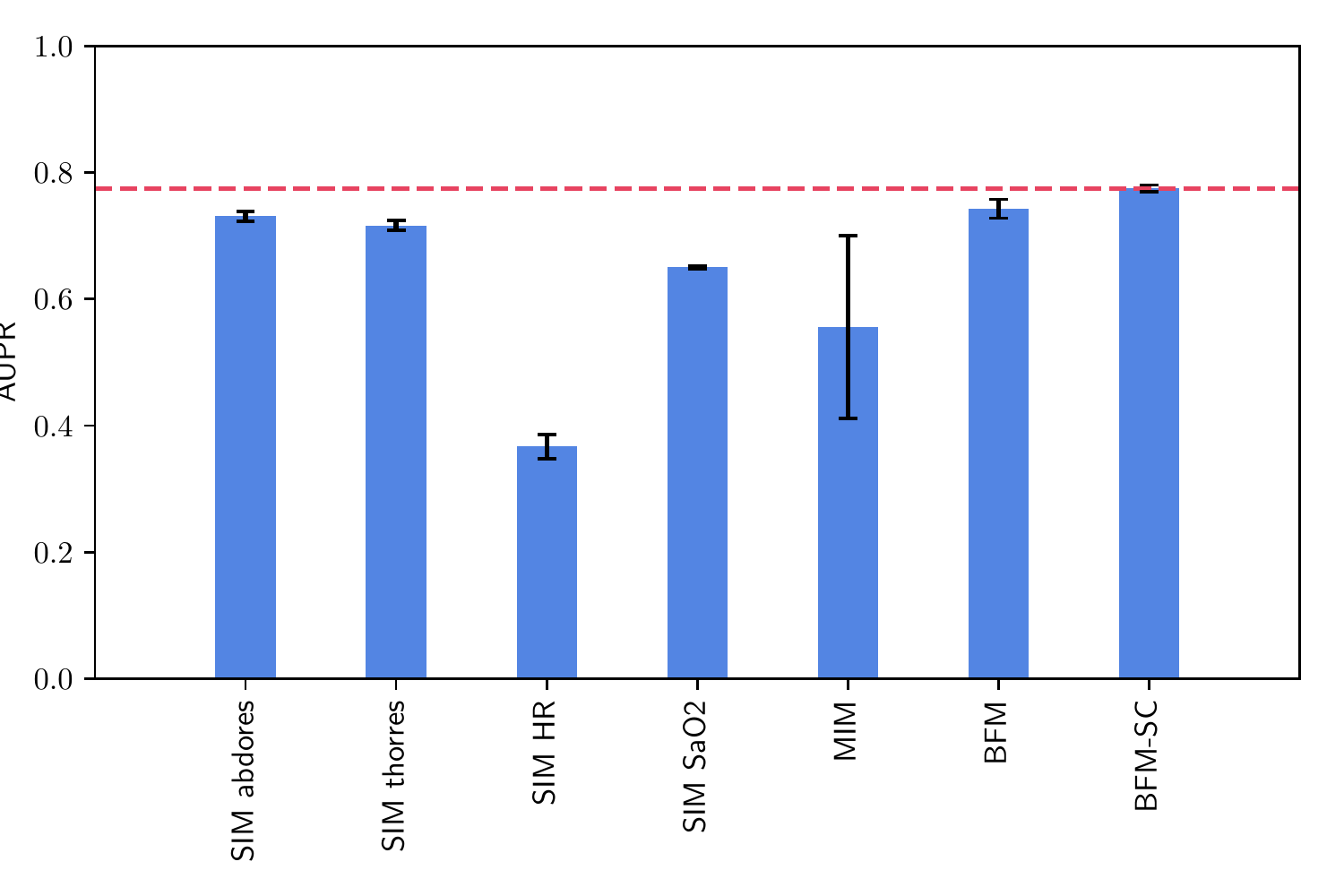}
		\caption{LSTM-based model.}
		\label{fig:}
	\end{subfigure}
	\caption{Summary of the robustness of the methods assessed by training and evaluating each configuration five times on the same dataset. The red line indicates the mean score of the BFM-SC.}
	\label{fig:same}
\end{figure}

The robustness of the methods, analyzed by training and evaluating the different configurations five times on the first group of patients is summarized in Figure~\ref{fig:same}. Most of the configurations have robust performance except for the BFM with the CNN-based model and the MIM with the LSTM-based model. This demonstrates the impact of the different underlying base models.

The BFM-SC results in a consistent increase in performance over the single input models as well as over the other sensor fusion configurations. Statistical significance is assessed using the T-test and Table~\ref{table:pvalues} demonstrates the resulting p-values for comparing the AUPR scores of each of the configurations to the AUPR scores of the BFM-SC configuration. For each experiment, the result is a $p<0.05$, demonstrating a statistically significant improvement of the BFM-SC over the other approaches.

\begin{table}[!htb]
	\centering
	\caption{Resulting p-values of the statistical significance tests using a paired T-test for performance analysis and a heteroscedastic T-test for robustness analysis. All experiments demonstrate a significant improvement of the BFM-SC over the original approaches.\label{table:pvalues}}
	\begin{tabular}{@{}lcccc@{}}
		\toprule
		& \multicolumn{2}{c}{CNN}  & \multicolumn{2}{c}{LSTM} \\ \midrule
		& performance & robustness & performance & robustness \\ \cmidrule(l){2-5} 
		SIM abdores & 0.016       & 0.001      & 0.049       & 0.000      \\
		SIM thorres & 0.009       & 0.001      & 0.013       & 0.000      \\
		SIM hr      & 0.000       & 0.000      & 0.000       & 0.000      \\
		SIM SaO2    & 0.003       & 0.000      & 0.010       & 0.000      \\
		MIM         & 0.000       & 0.000      & 0.037       & 0.039      \\
		BFM         & 0.014       & 0.017      & 0.001       & 0.013      \\ \bottomrule
	\end{tabular}
\end{table}

\begin{figure}[!htb]
	\centering
	\begin{subfigure}[h]{0.44\textwidth}
		\includegraphics[width=\textwidth]{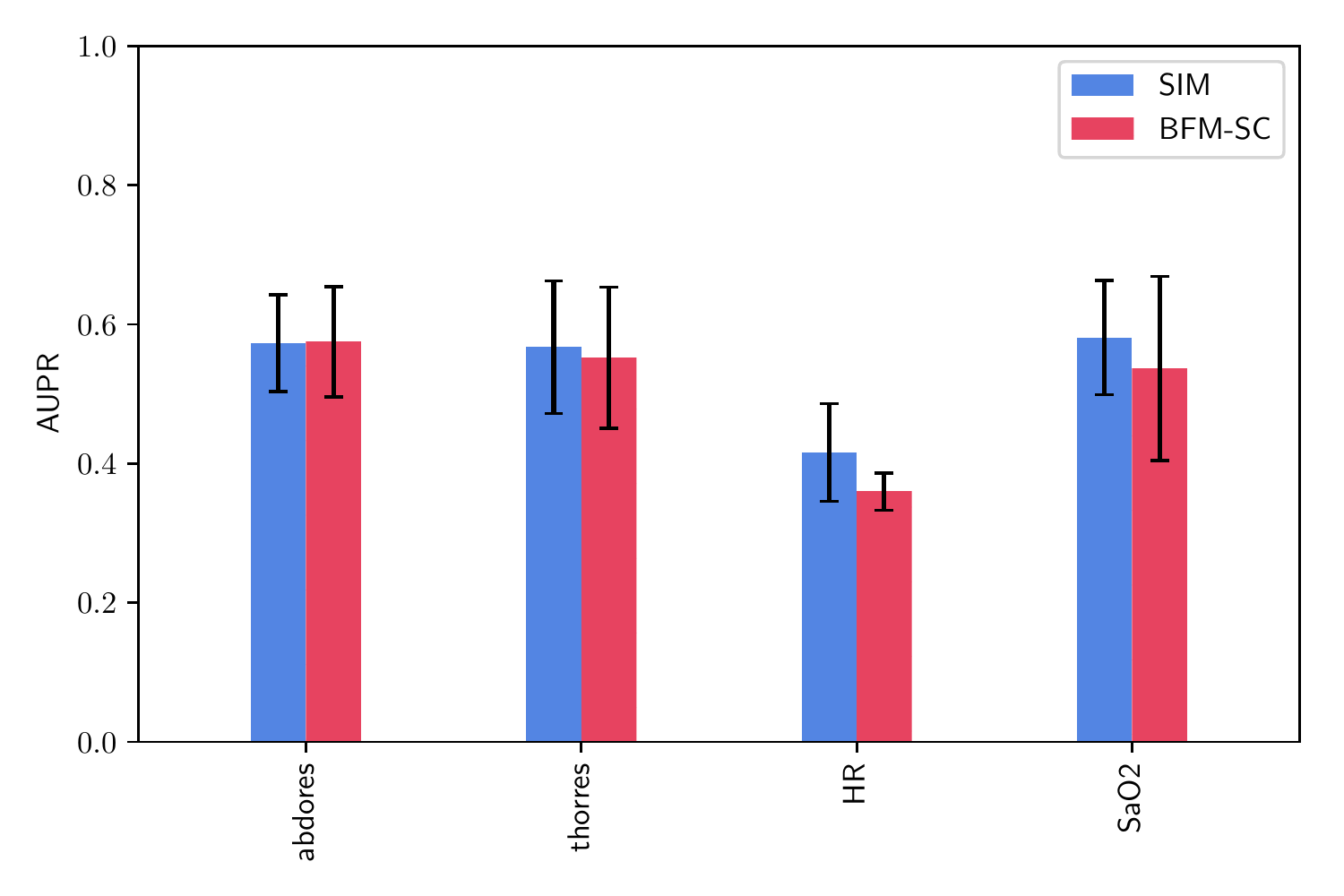}
		\caption{CNN-based model.}
		\label{fig:}
	\end{subfigure}
	~
	\begin{subfigure}[h]{0.44\textwidth}
		\includegraphics[width=\textwidth]{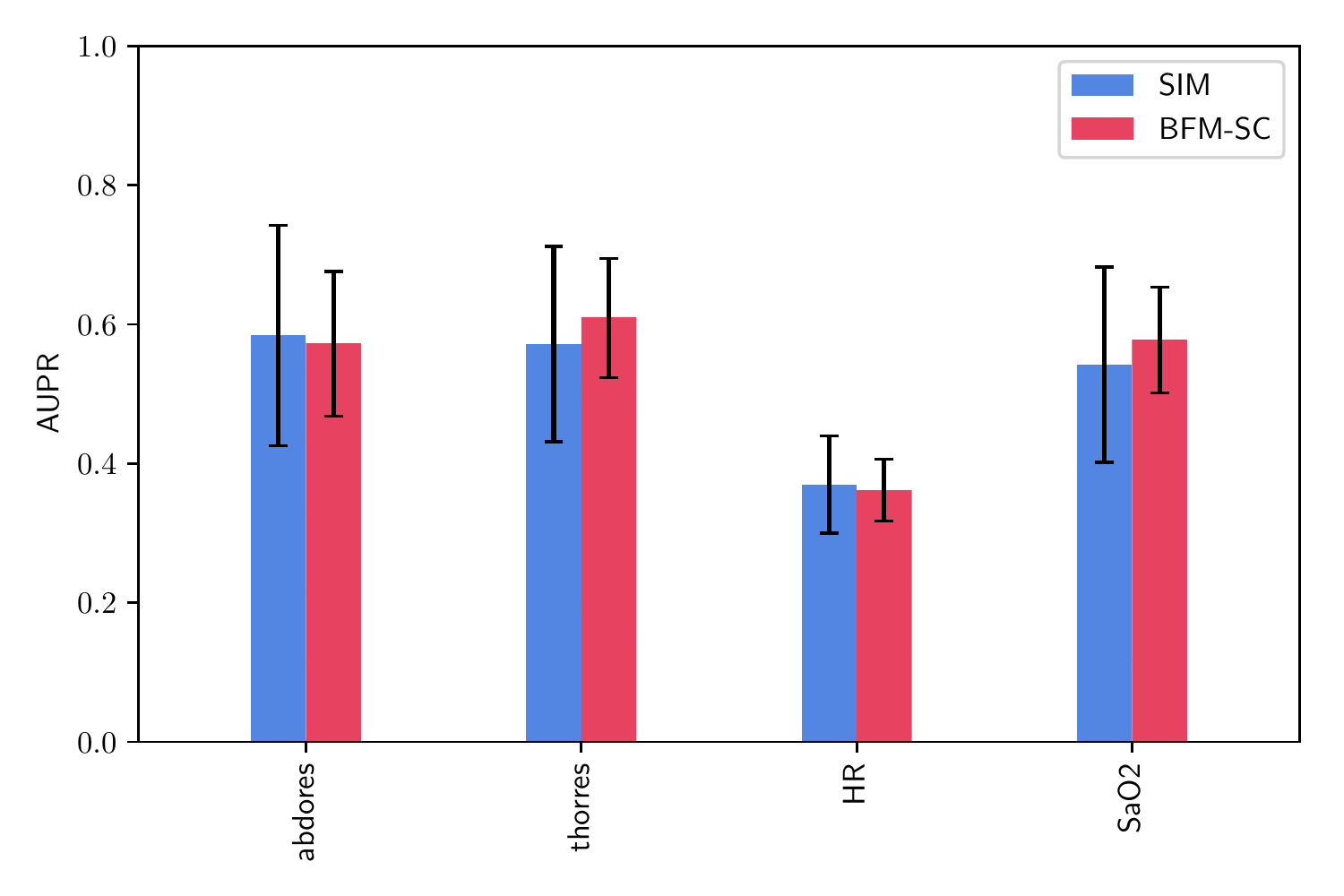}
		\caption{LSTM-based model.}
		\label{fig:}
	\end{subfigure}
	\caption{Performance of the individual branches in the BFM-SC for the prediction of sleep apnea events, compared to individual SIM models. The performance of each branch is close to the performance of a separately trained SIM model.}
	\label{fig:others}
\end{figure}

An added advantage of the BFM-SC approach is that it can be used to improve the interpretability of the results. Such interpretability and explainability is crucial to enable the deep learning methods to be used in clinical practice. The backward shortcut connections that are used during training can be removed at evaluation time. However, they can also be used to get a prediction per input sensor on top of the general fusion-based prediction. Figure~\ref{fig:others} demonstrates the performance of the individual branches in the BFM-SC setting for the detection of sleep apnea. The performance is close to that of a SIM model for each corresponding input. These extra outputs can lead to additional insights into the model decision and into which parameters are contributing to the detection.

There are some limitations to this study that warrant careful interpretation of the results. Although the performance of the BFM-SC and the other approaches was tested on five non-overlapping datasets, the analysis is still based on a single database. Validating these results on other sleep apnea databases could provide more weight to the posed conclusions. However, the SHHS-1 database is general and there is not a lot of variation in PSG measurements across other databases. The generalizability of the BFM-SC approach should also be tested on other use-cases than sleep apnea. In addition, the hyperparameters of the different models and configurations were based on the current state-of-the-art in literature. A complete analysis would require the hyperparameters to be optimized for each model, configuration and dataset. However, this would lead to a significant increase in computational cost. Initial experiments demonstrated that small changes to these parameters have limited impact on the results as the models have sufficient learning capacity in their current setting.

The proposed BFM-SC can be extended in several different ways. At the moment, each of the input signals is required to have the same sampling frequency. However, some signals are more informative at a higher sampling frequency whereas the learning process with other signals could be more efficient at lower frequencies. Further research should analyze methods of incorporating multi-frequency data into a BFM-SC setting. Moreover, the performance of the current configurations is impacted by the hard labeling strategy based on human annotations. The exact start and endpoints of an apnea event is not precise. Hence, further research should include smooth labeling. Finally, the current BFM-SC setup does not enable any data-level correlations to be learned. An interesting extension to the current work is the combination of data-level fusion and feature-level fusion into a single method.

\section{Conclusion}
\label{sec:conclusion}
Sleep apnea is typically diagnosed using multi-sensor data from a polysomnography device. As personnel is limited and waiting times are long, automated scoring methods are being developed. Deep learning algorithms have demonstrated good performance for sleep apnea detection in a single channel of respiratory data. However, these approaches ignore relevant and important markers in other signals. Various types of sensor fusion methods have been presented for medical use-cases and for other domains. In this work, the different types of sensor fusion were analyzed and a novel method based on backward shortcut connections was presented. It was demonstrated that the BFM-SC shows a significant and consistent improvement for automated sleep apnea detection in multi-sensor data. These results enable a more reliable automated sleep apnea detection method and present an opportunity for improving sensor fusion methods in other domains.

\section*{Acknowledgement}
This research received funding from the Flemish Government under the “Onderzoeksprogramma Artificiële Intelligentie (AI) Vlaanderen” programme.

\bibliographystyle{IEEEtran}
\bibliography{bibliography}

\end{document}